\def\year{2022}\relax
\documentclass[letterpaper,nocopyright]{article} 
\usepackage{aaai22}  
\usepackage{times}  
\usepackage{helvet}  
\usepackage{courier}  
\usepackage[hyphens]{url}  
\usepackage{graphicx} 
\usepackage{natbib}  
\usepackage{caption} 
\DeclareCaptionStyle{ruled}{labelfont=normalfont,labelsep=colon,strut=off} 
\usepackage{algorithm}
\usepackage{algorithmic}
\usepackage[center]{subfigure}
\usepackage{epstopdf}
\usepackage{amsmath}
\usepackage{txfonts}
\usepackage{amssymb}
\usepackage{newfloat}
\usepackage{listings}
\floatstyle{ruled}
\newfloat{listing}{tb}{lst}{}
\floatname{listing}{Listing}
\title{Multi-scale Graph Convolutional Networks with Self-Attention}
\author{
	Zhilong~Xiong \textsuperscript{},
	Jia~Cai \textsuperscript{\footnote{Corrsponding author}}
}
\affiliations{
    \textsuperscript{}School of Statistics and Mathematics, Guangdong University  of Finance $\&$ Economics\\
    Guangzhou,  510320, Guangdong, China.\\
   zhilongxiong98@outlook.com, jiacai1999@gdufe.edu.cn.\\
}

\usepackage{bibentry}
\usepackage{color}

\begin{document}

\maketitle

\begin{abstract}
Graph convolutional networks (GCNs) have achieved remarkable learning ability for dealing with various graph structural data recently. In general,  deep GCNs do not work well since graph convolution in conventional GCNs is a special form of Laplacian smoothing, which makes the representation of different nodes indistinguishable. In the literature, multi-scale information was employed in GCNs to enhance the expressive power of GCNs. However, over-smoothing phenomenon as a crucial issue of GCNs remains to be solved and investigated. In this paper, we propose two novel multi-scale GCN frameworks by incorporating self-attention mechanism and multi-scale information into the design of GCNs. Our methods greatly improve the computational efficiency and prediction accuracy of the GCNs model. Extensive experiments on both node classification and graph classification demonstrate the effectiveness over several state-of-the-art GCNs. Notably, the proposed two architectures can efficiently mitigate the over-smoothing problem of GCNs, and the layer of our model can even be increased to $64$.
\end{abstract}

\section{Introduction}\label{sec:intr}
Graph structural data related learning has drawn considerable attention recently. Graph neural networks (GNNs), particularly graph convolutional networks (GCNs) have been successfully utilized in  various fields of artificial intelligence, but not limited to  recommendation systems \cite{Sun2020AFF}, computer vision \cite{Casas2020SpAGNNSG}, molecular design \cite{2020stokes}, natural language processing \cite{Yao2019GraphCN}, node classification \cite{Kipf2017SemiSupervisedCW}, graph classification  \cite{xu2018how}, and clustering  \cite{Zhu2012MaxMarginNL}. Despite their great success, existing GCNs has low expressive power and may confront with the problem of over-smoothing, since graph convolution applies the same operation to all the neighbors of node. Therefore, almost all GCNs have shallow structures with only two or three layers. This  undoubtedly limits the expressive ability and the extraction ability of high-order neighbors of GCNs.

Several methods try to deal with the over-smoothing issue and deepen the networks.  \citet{Sun2019AdaGCNAG} developed  a novel AdaGCN by incorporating AdaBoost into the design of GCNs. JKNet \cite{xu2018representation}  used the information of each layer to improve the predictive ability of graph convolution. \citet{Rong2020dropedge} suggested removing a few edges of the graph randomly to mitigate the impact of over-smoothing,  whilst  \citet{luan2019break} utilized  multi-scale information to deepen GCNs.

However, the above-mentioned approaches yield steep increasing of computation  due to the accumulation of too many layers,  although the structure of GCNs is deepened. In this paper, we propose two general GCN frameworks by employing the idea of self-attention and multi-scale information. On one hand, the introduction of multi-scale information will enhance the expressive ability of GCNs by stacking many layers, which is scalable in depth. On the other hand, the self-attention mechanism will alleviate the over-smoothing problem, and reduce computational cost. Moreover, the proposed frameworks are flexible and can be built on any GCNs model. Extensive comparison against baselines in both node classification and graph classification is investigated. The experimental results indicate significant performance improvements on both graph and node classification tasks over a variety of real-world graph structural data.

\section{Related Work}\label{relaw}
In general, there are two convolution  operations  in the model of GCNs: spatial-based method and spectral-based approach. Spatial-based GCNs consider the aggregation method between the graph nodes. GAT \cite{Petar2018graph} used the attention mechanism to aggregate neighboring nodes on the graph, and GraphSAGE \cite{hamilton2017inductive}  utilized random walks to sample nodes and then aggregated them. Spetral-based GCNs  focus on redefining the convolution operation by utilizing Fourier transform \cite{Bruna2014SpectralNA} or wavelet transform \cite{Xu2019GraphWN} to define the graph signal in the spectral domain. However, the decomposition of  Laplacian matrix is too exhausted. Therefore, ChebyNet \cite{defferrard2016convolutional} was introduced, which employed Chebyshev polynomials to approximate the convolution kernel. Based upon ChebyNet, first-order Chebyshev polynomials \cite{Kipf2017SemiSupervisedCW} was used to approximate the convolution kernels to reduce computational cost. In addition, there are studies using ARMA filters \cite{bianchi2021graph} to approximate convolution kernels. \citet{he2021bernnet} proposed K-order Bernstein polynomials to approximate filters (BerNet).

In order to study deep GCNs, JKNet \cite{xu2018representation}  employed the information of front Layer to design deeper network.  \citet{Sun2019AdaGCNAG} used the Adaboost algorithm to design deep GCNs, whilst  multi-scale information was considered in \citet{luan2019break,liao2019lanczosnet}.

\section{Preliminaries}\label{main}

\textbf{Graph convolutional networks}: Denote $\mathcal{G} = (\mathcal{V},\mathcal{E})$ as an input undirected graph with $N$ nodes $v_i\in {\cal V}$, edges $(v_i,v_j)\in {\cal E}$. Let $A \in \mathbb{R}^{N\times N}$ be a symmetric adjacency matrix, and $D$ its corresponding diagonal degree matrix, i.e., $D_{ii}=\sum_j A_{ij}$. In conventional GCNs, the graph embedding of nodes with only one convolutional layer is depicted as
\begin{equation}\label{GCNexpre}
Z= {\rm ReLU}(\hat{A}XW_0)
\end{equation}
with $Z\in \mathbb{R}^{N\times K}$  the final embedding matrix (output logits) of  nodes before softmax, $K$ is the number of classes. Here $\hat{A} = \tilde{D}^{-\frac{1}{2} } \tilde{A}\tilde{D}^{-\frac{1}{2} }$ with  $\tilde{A}=A+I$ ($I$ stands for identity matrix),  $\tilde{D}$ is the degree matrix of $\tilde{A}$,  $X\in \mathbb{R}^{N\times d}$ is the  feature matrix with $d$ stands for the input dimension.  Furthermore, $W_0\in \mathbb{R}^{d\times H} $ denotes the input-to-hidden weight matrix for a hidden layer with $H$ feature maps.

To  design flexible deep GCNs for multi-scale distinct tasks, \citet{luan2019break} generalize vanilla GCNs in block Krylov subspace forms, which is defined by introducing a real analytic scalar function $g$, and rewrite $g(A)X$ as
\begin{eqnarray*}
	&&g(A) X\\
	&=& \sum^\infty _{n=0} \frac{g^{(n)}(0)}{n !}A^n X\\
	& =&[X, AX, \cdots, A^{m-1}X]\\
	&&[(\Gamma^{(\mathbb {S})}_0)^T, (\Gamma^{(\mathbb {S})}_1)^T,\cdots, (\Gamma^{(\mathbb {S})}_{m-1})^T],
\end{eqnarray*}
where $\mathbb {S}$ is a vector subspace of $\mathbb {R}^{d\times d}$ containing $I_d$ (the identity matrix),  $\Gamma^{(\mathbb {S})}_i \in \mathbb {R}^{d\times d}, i=1,\cdots, m-1$ are parameter matrix blocks. Thereafter, two architectures,  namely snowball and truncated Krylov networks are developed.

{\bf Snowball}, a densely-connected GCN, is depicted as the following
\begin{eqnarray*}
H_0 &= &X, \\
H_{\ell+1} & = &f(A[H_0, H_1,\cdots, H_\ell W_\ell]), \\
\quad && \ell=0,1,\cdots, m-1\\
{\bf C} &=&g([H_0, H_\ell,\cdots, H_n]W_n)\\
{\rm output} & = & {\rm softmax} (L^p {\bf C} W_C), p\in \{0,1\},
\end{eqnarray*}
where $W_\ell \in \mathbb {R}^{(\sum^\ell_{i=1}d_i)\times d_{\ell+1}}$, $W_n\in  \mathbb {R}^{(\sum^n_{i=1}d_i)\times d_C}$, $W_C\in  \mathbb {R}^{d_C\times d_0}$ are parameter matrices. $d_{\ell+1}$ is the number of output channels in the $\ell$-th layer, $f$ and $g$ stand for pointwise activation functions.

{\bf Truncated Krylov}, unlike snowball, will stack different scale information of $X, AX, \cdots, A^{m-1} X$ in each layer:
\begin{eqnarray*}
	H_0 &= &X, \\
	H_{\ell+1} &= & f([H_\ell, AH_\ell,\cdots, A^{m_\ell-1}H_\ell] W_\ell), \\
	\quad && \ell=0,1,\cdots, m-1\\
	{\bf C} &=&g(H_nW_n)\\
	{\rm output} & = & {\rm softmax} (L^p {\bf C} W_C), p\in \{0,1\},
\end{eqnarray*}
where  $W_\ell \in \mathbb {R}^{(m_\ell d_\ell)\times d_{\ell+1}}$, $W_n\in  \mathbb {R}^{d_n\times d_C}$ are parameter matrices.

From the iteration formula of snowball and truncated Krylov, we can observe that over-smoothing problem exist since no extra over-smoothing mitigation technique  is utilized. This will be justified in the experimental part.

\textbf{Self-attention}: Attention mechanisms is a widely used deep learning method, and has improved success when dealing with various tasks. Attention mechanism aims at focusing on important features, and less on unimportant features. Especially, self-attention \cite{vaswani2017attention} or intra-attention is an attention mechanism that allows the input features to interact with each other and find out which features they should focus on. Self-attention mechanism can be described mathematically as follows
\begin{equation}\label{self-att}
f= {\rm tanh}({\rm GNN}(X, A ))
\end{equation}
where  ${\rm GNN}(X, A )$ stands for the  graph convolution formula introduced by  \citet{Kipf2017SemiSupervisedCW}, i.e.,
\begin{equation}\label{GCNexpre}
H_{\ell+1} =  {\rm ReLU}(\hat{A}H_\ell W)
\end{equation}
with $H_0 = X$, $W\in  \mathbb {R}^{d_{h_\ell}\times d_{h_\ell}}$, where $d_{h_\ell}$ is the feature dimension of $H_\ell$. Here we utilize $\rm tanh$ activation function. In the sequel, we will observe that the introduction of self attention mechanism can reflect the topology as well as node features,  and extract the important features of multi-scale information. Thereby  improve the computational efficiency of the model, mitigate the over-smoothing issue.

\section{The proposed approaches}

\begin{figure*}[!htpb]
	\centering
	\subfigure[MGCN(H)]{\label{model(h)}\includegraphics[width=0.7\textwidth]{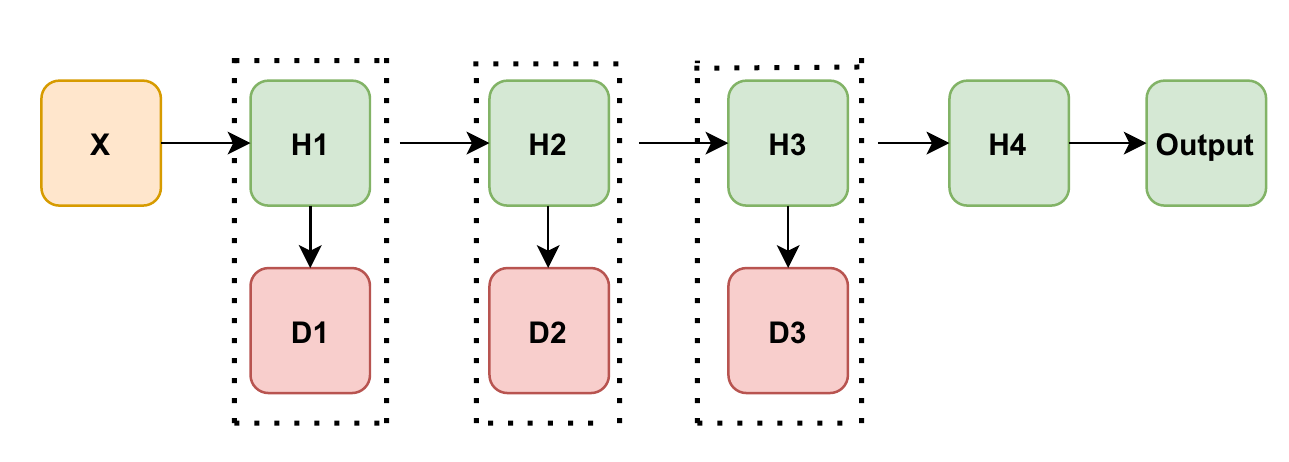}}
	\mbox{\hspace{0.1cm}}
	\subfigure[MGCN(G)]{\label{model(g)}\includegraphics[width=0.7\textwidth]{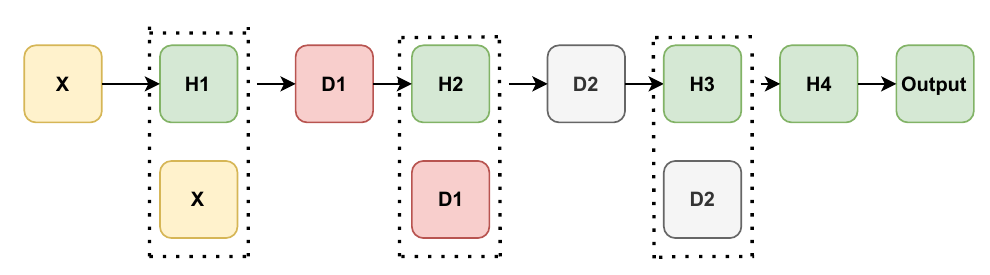}}
	\mbox{\hspace{0.1cm}}
	\vspace{-0.3cm}
	\caption{The hierarchical GCN architecture  and the global GCN architecture.  $H$ represents the result of computing the graph convolutional layer, and $D$ denotes the result of the calculation of the self-attention mechanism.}
	\label{model architecture}
\end{figure*}

As mentioned above,  multi-scale GCN may cause  over-smoothing and high computational burden.  Therefore, it is necessary to aggregate multi-scale information in an effective way for better predictions. Motivated by attention mechanisms, we propose two architectures: MGCN(H) and MGCN(G) to amend the deficiency of snowball and truncated Krylov methods. These methods concatenate multi-scale feature information by self-attention mechanism, which both have  the potential to be scaled to deeper architectures.

\textbf{MGCN(H)}: In order to allow the model to use the information of different scales without increasing the computational cost, motivated by the PeleeNet \cite{wang2018pelee}   designed for object detection, we develop a  multi-scale GCN (Fig. \ref{model(h)}), which adds an attention module in the middle of each layer. This module can expand the receptive field of the information achieved by the previous layer, concatenate the output of the previous layer and the  obtained information from the attention module to the subsequent layer. Hence, this kind of approach can not only enhance the feature ability of the GCNs, but also improve the prediction accuracy, and not increase the computational burden.  Let us address the detailed structure of it.  The first layer is the same as conventional GCN.
\begin{equation}
H_{1} = {\rm \sigma}(AXW_{0})
\end{equation}
where $W_{0}\in \mathbb{R}^{d\times d_{0} } $ is the input-to-hidden weight matrix, $\sigma$ is the activation function (e.g. $\rm tanh$ , $\rm ReLU$ ). The  attention module is computed as
\begin{equation}
D_{l} = {\rm tanh}({\rm GNN}(H_{l},A )),l=1,2,\cdots, n-1,
\end{equation}
\begin{equation}
C_{l} = {\rm Concate}(H_{l},D_{l}), l=1,2,\cdots, n-1,
\end{equation}
where $C_{l}$ are used as input to the next layer, $Concate$ stands for concatenate.   Hence, $C_\ell$ conducts concatenation operation. Similarly, we can even perform self-attention calculations on the feature matrix $X$, and the result will be combined  with itself as the input of the first layer. The  computation of graph convolution for the middle layers can be casted as
\begin{equation}
H_{l+1}  = {\rm \sigma}(AC_{l}W_{l}),l=1,2,...n-1.
\end{equation}
\begin{equation}
{\rm output}  =  {\rm softmax} (A H_{n} W_C)
\end{equation}
where  $W_l\in  \mathbb {R}^{(d_{\ell-1}+d_\ell)\times d_{l}}$ , $W_C\in  \mathbb {R}^{d_{n-1}\times d_C}$  are parameter matrices.  $d_{\ell}$ is the number of output channels in the $\ell$-th layer.

\textbf{MGCN(G)}: In \citet{luan2019break},  the snowball model transfers the information of previous layer to the next layer, so that the feature information of different scales could be well used,  and alleviate the  vanishing gradient problem, improve prediction accuracy. However,  it will cause a huge amount of calculation, and lead to the   over-smoothing problem. Therefore, we propose a self-attention based multi-scale global architecture   (Fig. \ref{model(g)}) that not only inherits the merits of snowball, but also overcome its shortcomings. The details are given as follows. The first layer is computed as
\begin{equation}
H_{1} = {\rm \sigma}(AXW_{0})
\end{equation}
where  $W_{0}\in \mathbb{R}^{d\times d_{0} } $ are parameter matrices.
\begin{equation}
C_{1} = {\rm Concate}(X,H_{1})
\end{equation}
\begin{equation}
D_{1} = {\rm tanh}({\rm GNN}(C_{1},A))
\end{equation}
The core idea for the computation of the subsequent layers is taking a self-attention module calculation after splicing the previous results, and  passing it to the next layer.
\begin{equation}
C_{l+1} = {\rm Concate}(H_{l+1},D_{l}),l=1,2,\cdots, n-1,
\end{equation}
\begin{equation}
D_{l+1} = {\rm tanh}({\rm GNN}(C_{l+1},A)),l=1,2,\cdots,n-1,
\end{equation}
\begin{equation}
H_{l+1} = {\rm \sigma}(AD_{l}W_{l}), l=1,2,\cdots, n-1,
\end{equation}
\begin{equation}
{\rm output}  =  {\rm softmax} (A H_{n} W_C)
\end{equation}
where  $W_l\in  \mathbb {R}^{(\sum^\ell_{i=0}d_i)\times d_{l}}$ , $W_C\in  \mathbb {R}^{d_{n-1}\times d_C}$ are parameter matrices.
Therefore, compared to snowball and truncated Krylov methods, the difference  lies in  that  extra attention  operations, i.e., Equations (5), (10) and (12),  are considered in the proposed two architectures.

\section{Experiments}\label{experi}
\subsection{Settings}
To validate the proposed global architecture and hierarchical architecture for graph representation learning, we evaluate our two multi-scale GCNs on both node classification and graph classification. All the experiments are performed under Ubuntu 16.04  in Inter CPU i6850K and 32 GB of RAM.

\subsection{Datasets}
We evaluate the  performance of the proposed  MGCN (H) and MGCN (G) on $6$ datasets with a large number of graphs ($>1k$) for the graph classification task, and $3$ commonly used citation networks for the semi-supervised node classification task. Details of the data statistics about the citation networks (graph datasets resp.) are  addressed in Table \ref{nodeclassdatasets} (Table \ref{graphclassdatasets} resp.)

\textbf{Citation networks}: The $3$ benchmark citation networks, namely,  Cora, Citeseer, and Pubmed \cite{Sen2008CollectiveCI} contains documents as node and citation links as directed edges, which stand for the citation relationships connected to documents.  The characteristics of nodes are representative words in documents, and the label rate here denotes the percentage of  node tags used for training.  We use undirected versions of the graphs for all the experiments although the networks are directed. In this paper, we use public splits \cite{yang2016revisiting} approach for training.

\textbf{Graph datasets}:  Among TU datasets \cite{2020TUDataset}, we select six datasets including D$\&$D, PROTEINS, MUTAGENICITY, NCI1, NCI109, and FRANKENSTEIN, where  D$\&$D, PROTEINS and MUTAGENICITY contain graphs of protein structures, whilst NCI1 and NCI109 are commonly used as benchmark datasets for graph classification.  FRANKENSTEIN is a set of molecular graphs  with node features containing continuous values.  For these $6$ datasets,  we randomly select  $80\%$,   $10\%$, and   $10\%$  of data for training,  validation, and test.

\begin{table}[!thb]
	\centering
\caption {Summary of the citation networks.}
	\begin{tabular}{lcccccc}
		\hline
		Dataset   &Cora&Citeseer&Pubmed& \\
		\hline
       Nodes   &2708&3327&19717\\
      Edges   &5429&4732&44338\\
      Features   &1433&3703&500\\
      Classes   &7&6&3\\
      Label Rate   &5.2\%&3.6\%&0.3\%\\
        \hline
	\end{tabular}
\label{nodeclassdatasets}
\end{table}

\begin{table*}[!thb]
	\centering
\caption {Summary of  the datasets used for graph classification.}
	\begin{tabular}{lcccccc}
		\hline
		Dataset   &D$\&$D &PROTEINS &NCI1 &NCI109 &FRANKENSTEIN &MUTAGENICITY\\
		\hline
      Number of Graphs   &1178    &1113     &4110  &4127    &4337     &4337  \\
      Avg. $\#$ of Nodes Per Graph   &284.32    &39.06     &29.87  &29.68    &16.90     &30.32   \\
      Avg. $\#$ of Edges Per Graph   &715.66    &72.82     &32.30  &32.13    &17.88     &30.77  \\
      Number of Classes   &2    &2     &2  &2    &2     &2   \\
        \hline
	\end{tabular}
\label{graphclassdatasets}
\end{table*}

\subsection{Comparison with state-of-the-art methods}
We compare against $3$ classical GCNs: graph convolutional network (GCN) \cite{Kipf2017SemiSupervisedCW}, graph attention network (GAT) \cite{Petar2018graph}, graph sample and aggregate (GraphSAGE) \cite{hamilton2017inductive}. Moreover, our model are based on multi-scale tactic and self-attention mechanism. Hence, classical  Chebyshev networks (Cheby) \cite{defferrard2016convolutional}, Self-Attention Graph Pooling (SAGPoolg and SAGPoolh) \cite{lee2019self} and two famous multi-scale deep convolutional networks: snowball  and truncated block Krylov network \cite{luan2019break}
are also considered as competitors.
\subsection{Parameter settings}

\textbf{Node classification}: For node classification task, we use RMSprop \cite{hinton2012neural} optimizer. There are several hyperparameters, including learning rate and weight decay. The learning rate takes values from the set $\{0.01, 4.7e-4, 4.7e-5\}$, whilst the weight decay is chosen from the set $\{0.03, 5e-3, 5e-4, 5e-5\}$. We choose  width of hidden layers in the set $\{200, 800, 1600, 1900, 3750\}$, number of hidden layers and dropout in the set $\{1,2,...,8\}$ and $\{0, 0.2\}$, respectively. To  achieve good training result, we utilize  adaptive number of episodes (but no more than 3000): the early stopping counter is set to be 100, the same as those described in  \citet{luan2019break}.
	
\textbf{Graph classification}: For graph classification task, Adam optimizer \cite{Kingma2015AdamAM} is employed.  We set learning rate $\gamma=5e-4$, pooling ratio equals to $1$, weight decay as $1e-4$, and dropout rate equals to $0.5$. The model considered in this paper has $128$ hidden units. We terminate the training if the validation loss does not improve for $50$ epochs (the maximum of epochs are set as $100K$). In addition, for the two proposed architectures, we use the MLP module in the last, and utilize the mean and the maximum readout, whilst the mean readout is employed for all the baselines.
\subsection{Experimental results}
As demonstrated in Fig. \ref{model architecture}, the MGCN (H) and the MGCN (G) stand for hierarchical GCN architecture and the global GCN architecture, respectively. We report the accuracy, and training time. The comparison results for node classification and graph classification are summarized in Tables \ref{accuracy_node} and \ref{accuracy_graph}, respectively.
\begin{table}[!thb]
	\centering
\caption {Average accuracy over $10$ runs for node classification.}
	\begin{tabular}{lcccccc}
		\hline
		Method   &Cora &Citeseer &Pubmed \\
		\hline
        Cheby   &78.0   &70.1    &69.8   \\
        GCN   &80.5    &68.7   &77.8   \\
        GAT   &83.0  &72.5  &79.0   \\
        GraphSAGE   &74.5   &67.2   &76.8   \\
        snowball   &83.6   &72.6    &79.5   \\
        truncated Krylov    &83.5   &73.9    &79.9\\
        \hline
        MGCN (H)   &83.4   &72.7   &79.5  \\
        MGCN (G)  &\textbf{84.0}  &\textbf{74.0}  &\textbf{80.0}  \\
        \hline
	\end{tabular}
\label{accuracy_node}
\end{table}

\begin{table*}[!thb]
	\centering
	\caption {Average accuracy over $10$ runs for graph classification.}
	\begin{tabular}{lcccccc}
		\hline
		Method   &D$\&$D &PROTEINS &NCI1 &NCI109 &FRANKENSTEIN &MUTAGENICITY\\
		\hline
		GCN   &73.26    &75.17   &76.29   &75.19    &62.70   &79.81\\
		GraphSAGE   &75.78  &74.01  &74.73   &74.17  &63.91  &78.75\\
		GAT   &77.30   &74.72   &74.90   &75.81   &59.90   &78.89\\
		SAGPool (g)   &76.19   &70.04    &74.18   &74.06   &62.57    &74.51\\
		SAGPool (h)   &76.45   &71.86    &67.45   &67.86   &61.73    &74.71\\
		snowball   &73.95   &71.43    &71.53   &78.02   &62.30    &83.68\\
		truncated Krylov   &79.54   &78.85    &\textbf{78.16}   &78.16   &78.16    &79.54\\
		\hline
		MGCN (H)   &82.53   &\textbf{82.07}   &78.10  &79.31   &74.48   &\textbf{81.15}\\
		MGCN (G)  &\textbf{83.91} &81.84    &77.24  &\textbf{80.23}  &\textbf{80.00}  &\textbf{81.15}\\
		\hline
	\end{tabular}
	\label{accuracy_graph}
\end{table*}
Experimental results in Tables \ref{accuracy_node} and \ref{accuracy_graph}  demonstrate significant improvement of MGCN model over the baselines. Specifically, for $3$ public citation networks and $6$ graph datasets, MGCN (H) or MGCN (G) achieves impressive improvement of the accuracy on  $3$ public citation  datasets, and outperforms the competitors on $4$ out of $6$ datasets.  Intuitively,  MGCN (H) and MGCN (G) are able to enhance the performance of GCNs, and provides better qualitative results for distinct types of graph structural  data.

\begin{figure*}[!htpb]
	\centering
	\subfigure[Accuracy on PROTEINS dataset with different layers]{\label{model(g)}\includegraphics[width=0.4\textwidth]{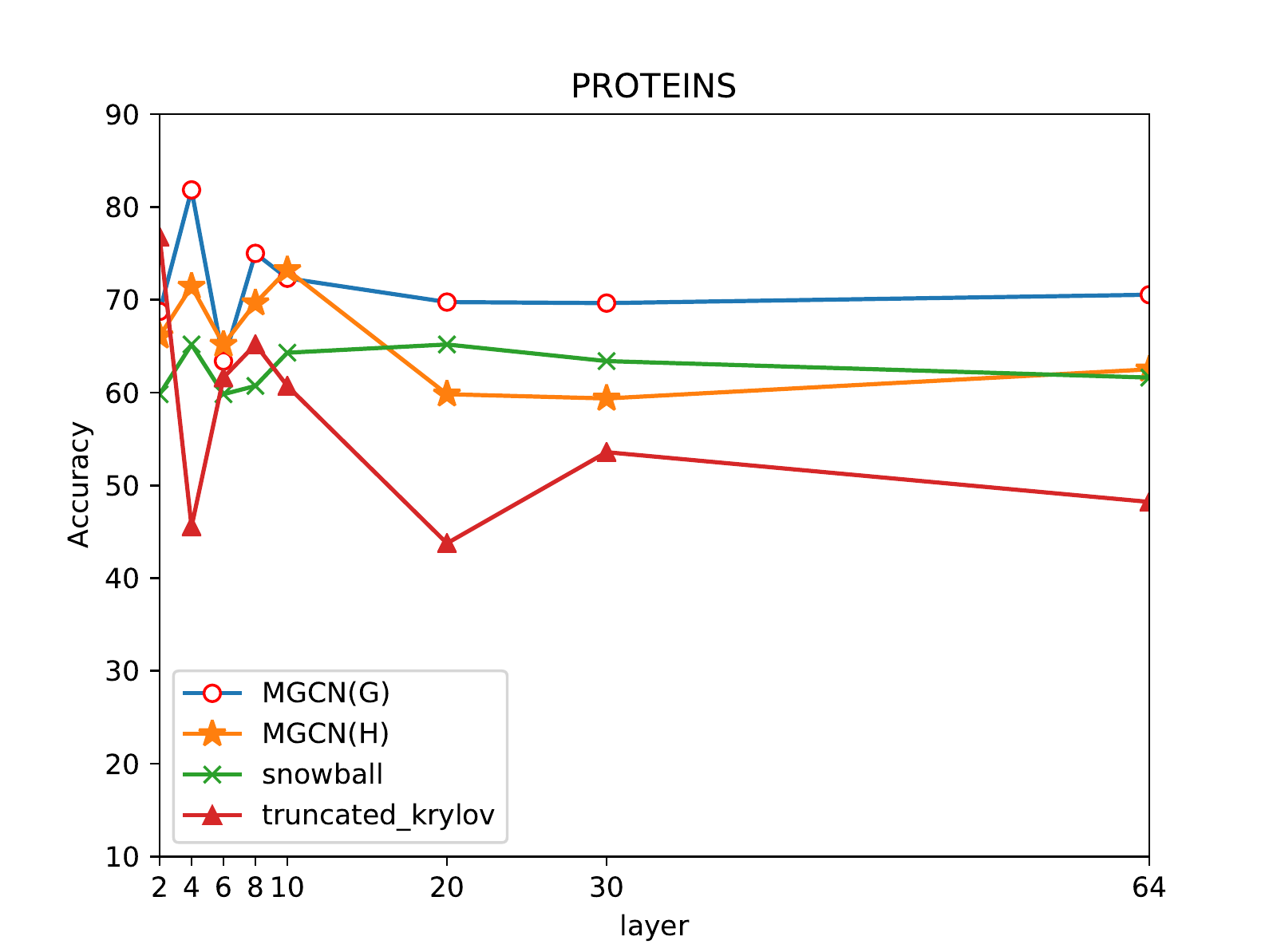}}
	\mbox{\hspace{0.1cm}}
	\subfigure[Accuracy on D$\&$D dataset with different layers]{\label{model(h)}\includegraphics[width=0.4\textwidth]{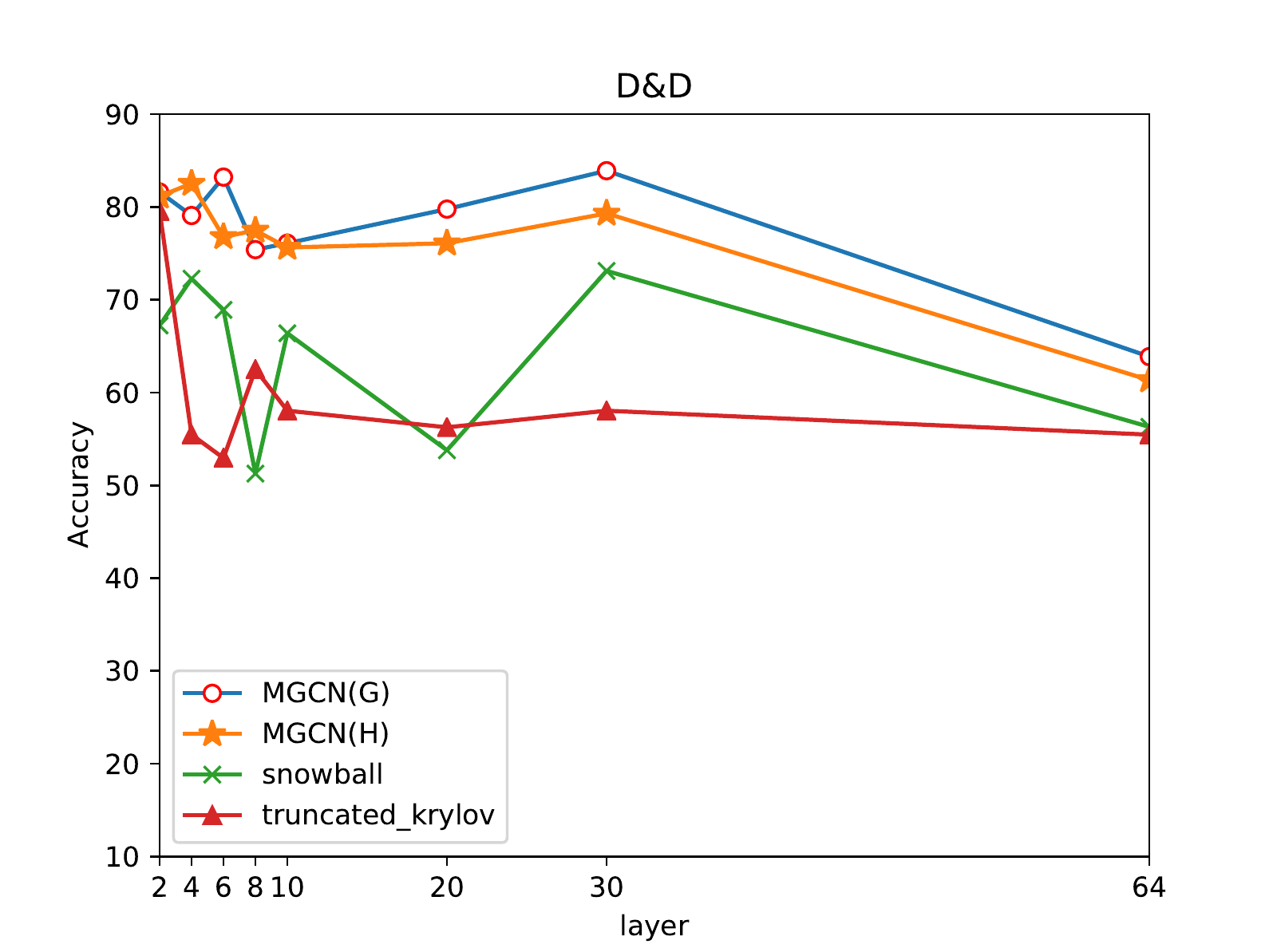}}
	\mbox{\hspace{0.1cm}}
	\vspace{-0.3cm}
	\caption{The accuracy on D$\&$D (a)  and PROTEINS (b) datasets with different layers.}
	\label{visual_display}
\end{figure*}

To demonstrate the effect of the proposed two architectures related with the over-smoothing issue, we also add experiments on the over-smoothing discussion. In Fig. \ref{visual_display}, Tables \ref{Oversmooth_MGCN(pro)} and \ref{Oversmooth_MGCN(dd)}, we can observe that the proposed MGCN (H) and MGCN (G) effectively ameliorate  the over-smoothing phenomenon and improve the accuracy. Even the number of layers are stacked to $30$ or $64$, the accuracy achieved is still the highest. However, the accuracy results of snowball and truncated Krylov methods drop rapidly when the depth of the network is increased.

Table \ref{time_cit} displays the comparison of training time between  MGCN (H),  MGCN (G) and two recently proposed multi-scale deep convolutional networks, namely, snowball and truncated Krylov.  We can see that MGCN (G)  runs faster on Cora and Citeseer datasets. We think the reason is that  the self-attention mechanism allow the model to focus on using multi-scale information,  instead of  accumulating the information from the previous layer to the next layer mechanically, which can achieve much less computational cost.

\begin{table*}[!ht]\normalsize
	\caption{Accuracy on PROTEINS dataset using  distinct number of layers.}
	\centering
	\begin{tabular}{ccccccccccc}
		\hline
		Method &2-layer  &4-layer  &6-layer  &8-layer &10-layer &20-layer &30-layer  &64-layer\\
		\hline
		snowball &59.82  &65.18  &59.82  &60.71 &64.29 &65.18 &63.39 &61.61 \\
		truncated Krylov &{\bf 76.78}  &45.54  &61.61  &65.18 &60.71 &43.75 &53.57  &48.21\\
		MGCN(H) &66.07  &71.43  &{\bf 65.18}  &69.64 &{\bf 73.21} &59.82 &59.36  &62.50\\
		MGCN(G) &68.75  &{\bf 81.84}  &63.39  &{\bf 75.00} &72.32 &{\bf 69.75} &{\bf 69.64} &{\bf 70.54}\\
		\hline
	\end{tabular}
	\label{Oversmooth_MGCN(pro)}
\end{table*}

\begin{table*}[!ht]\normalsize
	\caption{Accuracy on D$\&$D dataset using  distinct number of layers.}
	\centering
	\begin{tabular}{ccccccccccc}
		\hline
		Method &2-layer  &4-layer  &6-layer  &8-layer &10-layer &20-layer &30-layer  &64-layer\\
		\hline
		snowball &67.23  &72.27  &68.91  &51.26 &66.39 &53.78 &73.11  &56.30\\
		truncated Krylov &79.54  &55.46  &52.94  &62.50 &58.04 &56.25 &58.04 &55.46\\
		MGCN(H) &81.15  &{\bf 82.53}  &76.78  &{\bf 77.47} &75.63 &76.09 &79.31 &61.34\\
		MGCN(G) &{\bf 81.61}  &79.08  &{\bf 83.22}  &75.40 &{\bf 76.09} &{\bf 79.77} &{\bf 83.91}& {\bf  63.87}\\
		\hline
	\end{tabular}
	\label{Oversmooth_MGCN(dd)}
\end{table*}

\begin{table}[!h]
\caption{Average training time (s) comparison with 	snowball and truncated Krylov methods on $3$ citation networks.}
	\centering
	\begin{tabular}{lccc}
		\hline
		Method     &Cora  &Citeseer &Pubmed  \\
		\hline
		snowball	&484.10   &1228.57   &265.41    \\
		truncated Krylov   &213.81   &1194.94    &254.85   \\
		MGCN (H)   &\textbf{158.88}     &304.53    &219.92       \\
		MGCN (G)   &184.90     &\textbf{269.74}   &\textbf{184.60}       \\
		\hline
	\end{tabular}
	\label{time_cit}
\end{table}

\section{Conclusion and future work}\label{concl}

In this paper, we propose two novel multi-scale graph convolutional networks based on self-attention. Our methods can not only utilize multi-scale information to improve the expressive power of GCNs, but also alleviate the problems of large scale calculation,  and over-smoothing of the previous multi-scale graph convolution model. We validate the performance of the proposed frameworks on both node classification and graph classification tasks. Our framework is universal, and one can  replace the self-attention module via other graph pooling methods, for instance,  gPool \cite{gao2019graph},  SortPool \cite{zhang2018end}, DiffPool \cite{10.5555/3327345.3327389}, Set2Set \cite{Vinyals2016OrderMS} etc.

\section{Acknowledgement}
The work described in this paper was supported partially by the National Natural Science Foundation of China (11871167), Special Support Plan for High-Level Talents of Guangdong Province (2019TQ05X571), Foundation of Guangdong Educational Committee (2019KZDZX1023), Project of Guangdong Province Innovative Team (2020WCXTD011)

\bibliographystyle{aaai22}
\bibliography{aaai222}

\begin{thebibliography}{31}
\providecommand{\natexlab}[1]{#1}

\bibitem[{Bianchi et~al.(2021)Bianchi, Grattarola, Livi, and
  Alippi}]{bianchi2021graph}
Bianchi, F.~M.; Grattarola, D.; Livi, L.; and Alippi, C. 2021.
\newblock Graph neural networks with convolutional arma filters.
\newblock \emph{IEEE Transactions on Pattern Analysis and Machine
  Intelligence}.

\bibitem[{Bruna et~al.(2014)Bruna, Zaremba, Szlam, and
  Lecun}]{Bruna2014SpectralNA}
Bruna, J.; Zaremba, W.; Szlam, A.; and Lecun, Y. 2014.
\newblock Spectral Networks and Locally Connected Networks on Graphs.
\newblock In \emph{ICLR}.

\bibitem[{Casas et~al.(2020)Casas, Gulino, Liao, and
  Urtasun}]{Casas2020SpAGNNSG}
Casas, S.; Gulino, C.; Liao, R.; and Urtasun, R. 2020.
\newblock SpAGNN: Spatially-Aware Graph Neural Networks for Relational Behavior
  Forecasting from Sensor Data.
\newblock \emph{2020 IEEE International Conference on Robotics and Automation
  (ICRA)}, 9491--9497.

\bibitem[{Defferrard, Bresson, and
  Vandergheynst(2016)}]{defferrard2016convolutional}
Defferrard, M.; Bresson, X.; and Vandergheynst, P. 2016.
\newblock Convolutional neural networks on graphs with fast localized spectral
  filtering.
\newblock \emph{NeurIPS}, 29: 3844--3852.

\bibitem[{Gao and Ji(2019)}]{gao2019graph}
Gao, H.; and Ji, S. 2019.
\newblock Graph u-nets.
\newblock In \emph{ICML}, 2083--2092.

\bibitem[{Hamilton, Ying, and Leskovec(2017)}]{hamilton2017inductive}
Hamilton, W.~L.; Ying, R.; and Leskovec, J. 2017.
\newblock Inductive representation learning on large graphs.
\newblock In \emph{NeurIPS}, 1025--1035.

\bibitem[{He et~al.(2021)He, Wei, Huang, and Xu}]{he2021bernnet}
He, M.; Wei, Z.; Huang, Z.; and Xu, H. 2021.
\newblock BernNet: Learning Arbitrary Graph Spectral Filters via Bernstein
  Approximation.
\newblock \emph{arXiv preprint arXiv:2106.10994}.

\bibitem[{Hinton, Srivastava, and Swersky(2012)}]{hinton2012neural}
Hinton, G.; Srivastava, N.; and Swersky, K. 2012.
\newblock Neural networks for machine learning lecture 6a overview of
  mini-batch gradient descent.
\newblock \emph{Cited on}, 14(8): 2.

\bibitem[{Kingma and Ba(2015)}]{Kingma2015AdamAM}
Kingma, D.; and Ba, J. 2015.
\newblock Adam: A Method for Stochastic Optimization.
\newblock In \emph{ICLR}.

\bibitem[{Kipf and Welling(2017)}]{Kipf2017SemiSupervisedCW}
Kipf, T.; and Welling, M. 2017.
\newblock Semi-Supervised Classification with Graph Convolutional Networks.
\newblock In \emph{ICLR}.

\bibitem[{Lee, Lee, and Kang(2019)}]{lee2019self}
Lee, J.; Lee, I.; and Kang, J. 2019.
\newblock Self-attention graph pooling.
\newblock In \emph{ICML}, 3734--3743.

\bibitem[{Liao et~al.(2019)Liao, Zhao, Urtasun, and Zemel}]{liao2019lanczosnet}
Liao, R.; Zhao, Z.; Urtasun, R.; and Zemel, R. 2019.
\newblock LanczosNet: Multi-Scale Deep Graph Convolutional Networks.
\newblock In \emph{ICLR}.

\bibitem[{Luan et~al.(2019)Luan, Zhao, Chang, and Precup}]{luan2019break}
Luan, S.; Zhao, M.; Chang, X.-W.; and Precup, D. 2019.
\newblock Break the Ceiling: Stronger Multi-scale Deep Graph Convolutional
  Networks.
\newblock In \emph{NeurIPS}, volume~32.

\bibitem[{Morris et~al.(2020)Morris, Kriege, Bause, Kersting, and
  Neumann}]{2020TUDataset}
Morris, C.; Kriege, N.~M.; Bause, F.; Kersting, K.; and Neumann, M. 2020.
\newblock TUDataset: A collection of benchmark datasets for learning with
  graphs.

\bibitem[{Rong et~al.(2020)Rong, Huang, Xu, and Huang}]{Rong2020dropedge}
Rong, Y.; Huang, W.; Xu, T.; and Huang, J. 2020.
\newblock DropEdge: Towards Deep Graph Convolutional Networks on Node
  Classification.
\newblock In \emph{ICLR}.

\bibitem[{Sen et~al.(2008)Sen, Namata, Bilgic, Getoor, Gallagher, and
  Eliassi-Rad}]{Sen2008CollectiveCI}
Sen, P.; Namata, G.; Bilgic, M.; Getoor, L.; Gallagher, B.; and Eliassi-Rad, T.
  2008.
\newblock Collective Classification in Network Data.
\newblock \emph{AI Mag.}, 29: 93--106.

\bibitem[{Stokes et~al.(2020)Stokes, Yang, Swanson, Jin, and
  Collins}]{2020stokes}
Stokes, J.~M.; Yang, K.; Swanson, K.; Jin, W.; and Collins, J.~J. 2020.
\newblock A Deep Learning Approach to Antibiotic Discovery.
\newblock \emph{Cell}, 180(4): 688--702.e13.

\bibitem[{Sun et~al.(2020)Sun, Guo, Zhang, Zhang, Regol, Hu, Guo, Tang, Yuan,
  He, and Coates}]{Sun2020AFF}
Sun, J.; Guo, W.; Zhang, D.; Zhang, Y.; Regol, F.; Hu, Y.; Guo, H.; Tang, R.;
  Yuan, H.; He, X.; and Coates, M. 2020.
\newblock A Framework for Recommending Accurate and Diverse Items Using
  Bayesian Graph Convolutional Neural Networks.
\newblock \emph{Proceedings of the 26th ACM SIGKDD International Conference on
  Knowledge Discovery \& Data Mining}.

\bibitem[{Sun, Lin, and Zhu(2021)}]{Sun2019AdaGCNAG}
Sun, K.; Lin, Z.; and Zhu, Z. 2021.
\newblock AdaGCN: Adaboosting Graph Convolutional Networks into Deep Models.
\newblock In \emph{ICLR}.

\bibitem[{Vaswani et~al.(2017)Vaswani, Shazeer, Parmar, Uszkoreit, Jones,
  Gomez, Kaiser, and Polosukhin}]{vaswani2017attention}
Vaswani, A.; Shazeer, N.; Parmar, N.; Uszkoreit, J.; Jones, L.; Gomez, A.~N.;
  Kaiser, {\L}.; and Polosukhin, I. 2017.
\newblock Attention is all you need.
\newblock In \emph{NeurIPS}, 5998--6008.

\bibitem[{Veli{\v{c}}kovi{\'c} et~al.(2018)Veli{\v{c}}kovi{\'c}, Cucurull,
  Casanova, Romero, Li$\grave{o}$, and Bengio}]{Petar2018graph}
Veli{\v{c}}kovi{\'c}, P.; Cucurull, G.; Casanova, A.; Romero, A.;
  Li$\grave{o}$, P.; and Bengio, Y. 2018.
\newblock Graph Attention Networks.
\newblock In \emph{ICLR}.

\bibitem[{Vinyals, Bengio, and Kudlur(2016)}]{Vinyals2016OrderMS}
Vinyals, O.; Bengio, S.; and Kudlur, M. 2016.
\newblock Order Matters: Sequence to sequence for sets.
\newblock In \emph{ICLR}.

\bibitem[{Wang et~al.(2018)Wang, Li, Ao, and Ling}]{wang2018pelee}
Wang, R.; Li, X.; Ao, S.; and Ling, C. 2018.
\newblock Pelee: A Real-Time Object Detection System on Mobile Devices.
\newblock In \emph{NeurIPS}.

\bibitem[{Xu et~al.(2019{\natexlab{a}})Xu, Shen, Cao, Qiu, and
  Cheng}]{Xu2019GraphWN}
Xu, B.; Shen, H.; Cao, Q.; Qiu, Y.; and Cheng, X. 2019{\natexlab{a}}.
\newblock Graph Wavelet Neural Network.
\newblock In \emph{ICLR}.

\bibitem[{Xu et~al.(2019{\natexlab{b}})Xu, Hu, Leskovec, and
  Jegelka}]{xu2018how}
Xu, K.; Hu, W.; Leskovec, J.; and Jegelka, S. 2019{\natexlab{b}}.
\newblock How Powerful are Graph Neural Networks?
\newblock In \emph{ICLR}.

\bibitem[{Xu et~al.(2018)Xu, Li, Tian, Sonobe, Kawarabayashi, and
  Jegelka}]{xu2018representation}
Xu, K.; Li, C.; Tian, Y.; Sonobe, T.; Kawarabayashi, K.-i.; and Jegelka, S.
  2018.
\newblock Representation learning on graphs with jumping knowledge networks.
\newblock In \emph{ICML}, 5453--5462.

\bibitem[{Yang, Cohen, and Salakhudinov(2016)}]{yang2016revisiting}
Yang, Z.; Cohen, W.; and Salakhudinov, R. 2016.
\newblock Revisiting semi-supervised learning with graph embeddings.
\newblock In \emph{ICML}, 40--48.

\bibitem[{Yao, Mao, and Luo(2019)}]{Yao2019GraphCN}
Yao, L.; Mao, C.; and Luo, Y. 2019.
\newblock Graph Convolutional Networks for Text Classification.
\newblock In \emph{AAAI}.

\bibitem[{Ying et~al.(2018)Ying, You, Morris, Ren, Hamilton, and
  Leskovec}]{10.5555/3327345.3327389}
Ying, R.; You, J.; Morris, C.; Ren, X.; Hamilton, W.~L.; and Leskovec, J. 2018.
\newblock Hierarchical Graph Representation Learning with Differentiable
  Pooling.

\bibitem[{Zhang et~al.(2018)Zhang, Cui, Neumann, and Chen}]{zhang2018end}
Zhang, M.; Cui, Z.; Neumann, M.; and Chen, Y. 2018.
\newblock An end-to-end deep learning architecture for graph classification.
\newblock In \emph{AAAI}.

\bibitem[{Zhu(2012)}]{Zhu2012MaxMarginNL}
Zhu, J. 2012.
\newblock Max-Margin Nonparametric Latent Feature Models for Link Prediction.
\newblock In \emph{ICML}.

\end{thebibliography}

\end{document}